\documentclass{article}

\usepackage{PRIMEarxiv}
\usepackage{amsmath}

\usepackage[utf8]{inputenc} 
\usepackage[T1]{fontenc}    
\usepackage{hyperref}       
\usepackage{url}            
\usepackage{booktabs}       
\usepackage{amsfonts}       
\usepackage{nicefrac}       
\usepackage{microtype}      
\usepackage{lipsum}
\usepackage{fancyhdr}       
\usepackage{graphicx}       
\graphicspath{{media/}}     
\usepackage{amsmath} 

\title{Sensitivity Analysis on Loss Landscape
}

\author{
  Salman Faroz \\ 
  \texttt{stsfaroz@gmail.com} \\
}

\begin{document}
\maketitle

\begin{abstract}
Gradients can be employed for sensitivity analysis. Here, we leverage the advantages of the Loss Landscape to comprehend which independent variables impact the dependent variable. We seek to grasp the loss landscape by utilizing first, second, and third derivatives through automatic differentiation. we know  that Spearman's rank correlation coefficient can detect the monotonic relationship between two variables. However, I have found that second-order gradients, with certain configurations and parameters, provide information that can be visualized similarly to Spearman results,In this approach, we incorporate a loss function with an activation function, resulting in a non-linear pattern. Each exploration of the loss landscape through retraining yields new valuable information. Furthermore, the first and third derivatives are also beneficial, as they indicate the extent to which independent variables influence the dependent variable.
\end{abstract}
\section{Introduction}
The sensitivity analysis of a data through a focused exploration of the multiple parts of a loss landscape. By exploring the complex link that exists between the model's gradients and loss values, and by using the first three derivatives in particular, one may ascertain how these gradients change in response to different loss. This method offers a more thorough knowledge of the rate of change of the loss landscape across various gradients than traditional sensitivity analyses do. The goal of the paper is to provide a comprehensive representation of the loss landscape and highlight the importance of sensitivity analysis in supporting machine learning model robustness by exposing the theoretical underpinnings, experimental results, and wider implications of sensitivity analysis, going beyond conventional practices. 

we use straightforward neural network architecture with a fixed configuration that relies on specific choices of activation functions. This unique architecture is chosen deliberately due to its proven effectiveness in achieving the desired outcomes. The emphasis on the selection of these activation functions is crucial, as they play a pivotal role in shaping the network's behavior and contributing to the overall success of the model.

In summary, this study presents a neural network architecture with a fixed configuration, leveraging specific activation functions to unravel the intricacies of the loss landscape. By employing the first three derivatives, with a notable emphasis on second derivatives, and visualizing an example of gradients, we offer a comprehensive exploration of sensitivity analysis by plotting. 

\section{Related Work}
Conducting sensitivity analysis proves to be a formidable task, especially with non-linear and extensive datasets.This paper \cite{sanchez2021uncertainty} investigates the mathematical foundations of sensitivity theory using concepts of nonlinear functional analysis, formulating a more general sensitivity theory for physical problems characterized by systems of nonlinear equations and nonlinear functions as responses.

Even in scenarios where causation is evident, and lurking variables are absent, extracting precise information about the impact of independent variables on the dependent variable remains challenging due to the limitations of conventional statistical methods. Prior research has predominantly centered on the final weights of neural networks or the application of Jacobian/Hessian matrices.

A gradient-based search algorithm for feature selection \cite{sheth2019feature}. Their approach extends a recent result on the estimation of learnability in the sublinear data regime by showing that the calculation can be performed iteratively (i.e., in mini-batches) and in linear time and space with respect to both the number of features D and the sample size N. Another feature-selection method workflow that leverages a gradient boosting framework and statistical feature analyses to identify a subset of features, in a recursive manner, which maximizes their relevance to the target variable or classes \cite{jung2023gradient}.There has been extensive research on sensitivity analysis, including a tutorial on its application in clinical trials \cite{thabane2013tutorial}, an introduction to its principles and methods \cite{iooss2016introduction}, and a systematic review of its practices \cite{saltelli2023why}.

In our distinctive approach, we deviate by incorporating the complete gradient history, moving beyond a singular focus on final weights. By adhering to the first principles of derivatives, our methodology not only facilitates a clearer understanding but also augments analytical power. This departure from conventional practices offers an intuitive exploration of sensitivity analysis, particularly in the intricate landscape of non-linear and expansive datasets. Through our emphasis on the overall gradient history, we contribute to a more nuanced understanding of the relationships between independent and dependent variables, addressing the challenges posed by complex datasets

\section{Model Setup}

The model's configuration involves a series of components, employing gradient descent for optimization. Automatic differentiation facilitates backpropagation, allowing for the storage of the first three derivatives and the loss to conduct sensitivity analysis in subsequent analyses. Automatic differentiation plays a pivotal role in the backpropagation process by efficiently computing gradients. This computational efficiency is particularly valuable when dealing with intricate models and large datasets. The stored derivatives and loss information are integral not only for sensitivity analysis, but also for diagnosing the model's performance and identifying potential areas of improvement.

\subsection{Lean Network Design}  
In using this simple neural network, we're exploring why we opt for simplicity. Computing gradients is expensive, especially when calculating the first three derivatives. Despite its simplicity, this network can handle any dataset, and our primary goal isn't necessarily reaching the absolute minimum. Instead, we aim to minimize the loss reasonably. The simplicity of this network provides a clearer intuitive understanding of neural networks. Even though it's straightforward, it can still have multiple local minima.

\begin{equation}
\text{Input layer: } \mathbf{x} \in \mathbb{R}^n \hspace{0.2cm} 
\text{Output layer: } \mathbf{z} \in \mathbb{R}^1
\end{equation}

In this setup, the input layer is represented by the vector x in a space of n-dimensional real numbers, while the output layer is denoted by z as a single real number. This straightforward design, focusing on simplicity, captures n features in the input layer and produces a single output in the output layer. This uncomplicated structure not only ensures computational efficiency but also adheres to the principle of simplicity in neural network design.

\subsection{Activation Functions: The Engine of Operation}
The choice of activation function plays a crucial role in the functioning of neural network. After conducting experiments, it became evident that different activation functions, such as relu and others, significantly impact both the loss value and its gradient. For instance, relu confines the outputs. The functions $f(x) , f'(x)$ of the activation function have a profound effect on the landscape of the loss function. It's essential to note that changing the activation function alters the shape of the loss surface. While some activation functions have proven effective for learning, it's equally crucial to consider their limitations. Despite their proficiency, unexpected behaviors in gradients and the loss landscape can lead to unpredictable outcomes.
Furthermore, we are considering the hyperbolic tangent activation function 

It's noteworthy that the function tanh itself is not confined within this range, highlighting the nuanced behavior of the tanh activation function. Nevertheless, after careful evaluation, I have chosen the hyperbolic tangent activation function tanh as our primary activation function in our model due to its specific characteristics and suitability for our objectives.Adding to the considerations.it's worth noting that even though the hyperbolic tangent activation function has a derivative range of 0-1, the function $f(x)$ itself is not constrained within this range.

The options are significantly more varied than tanh in the complex world of activation functions. Take into account the Swish activation function, a self-gating function that adds non-linearity and frequently outperforms conventional options like ReLU in specific situations. Although its smooth gradient properties help to ensure steady training, its higher computing requirements call for careful attention. Although these activation functions have desirable qualities, applying them needs careful consideration of the particulars of the dataset as well as the goals of the neural network model. Finding the ideal activation function is a never-ending process, and knowing the subtle differences between them is quite helpful.

\subsection{Navigating through automatic differentiation} 
Automatic differentiation (AD) is a set of techniques used to compute the derivative of a function specified by a computer program. It exploits the fact that every computer calculation, no matter how complicated, executes a sequence of elementary arithmetic operations and functions. By repeatedly applying the chain rule (\ref{chainrule}) to these operations, AD can compute partial derivatives of arbitrary order accurately.

\begin{equation}
\label{chainrule}
\frac{\partial L}{\partial w_{ij}} = \sum_k \frac{\partial L}{\partial y_k} \frac{\partial y_k}{\partial z_k} \frac{\partial z_k}{\partial w_{ij}}
\end{equation}

where:
- $L$ is the loss function,
- $w_{ij}$ represents the weights of the neural network,
- $y_k$ is the output of the $k$-th neuron in the output layer,
- $z_k$ is the weighted sum of the inputs to the $k$-th neuron in the output layer.

Throughout this exploration, I systematically recorded key derivatives—first-order, second, and third-order—enhancing our understanding of the neural network's unique dynamics. This automatic differentiation provides valuable insights and guiding refined optimization strategies. I also kept an eye on epoch-wise loss values with gradients, giving us a dynamic view of the model's learning progress.

\section{Loss Landscape Perspective}
Loss landscape is changes as we change the loss function so we stick at more reliable function , so the choice is Mean Squared Error, and loss surface also changes by activation function so as I mentioned above activation function is chosen by experiment , even though as we can imagine them intuitively the surface affected by from multiple parameters such as weights so we have sticking with Xavier glorot initialization (\ref{glorot}), the choice of Normal distribution is very important since we want to explore the loss landscape a lot by every time we retrain it , but the uniform distribution will limit the exploration a bit , so we are gonna stick with normal. The structure of neural loss functions and their landscapes significantly influence the generalization of neural networks. also that certain network architectures and training parameters can shape these landscapes and the minimizers found within them.\cite{li2018visualizing}

\begin{equation}
\label{glorot}
\text w_{ij} \sim \mathcal{N}\left(0, \frac{2}{n_{\text{in}} + n_{\text{out}}}\right) ; \\
y = \tanh(w_{ij}x + b) ; \\
L = \frac{1}{n}\sum_{i=1}^{n}(y_i - t_i)^2 \\
\end{equation}

\begin{figure}
    \centering
    \includegraphics[width=0.5\linewidth]{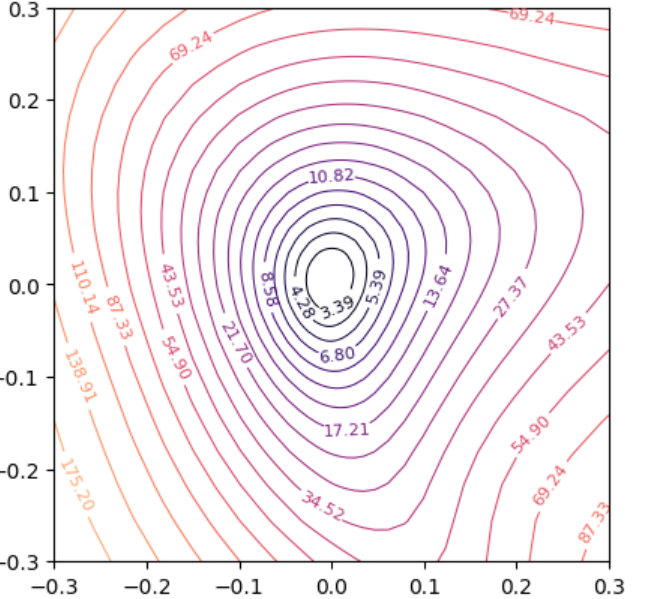}
    \caption{Loss Surface Of Experimented Dataset}
    \label{fig:loss}
\end{figure}

\paragraph{Fixed Configuration's Impact on Loss Dynamics} The configuration of the loss surface in machine learning is notably unpredictable. When observing the 2D representation of the loss surface \cite{hol2020visualizing} in the Figure \ref{fig:loss} above, which illustrates the dataset's loss variation, it becomes even more challenging to ascertain the presence of multiple local minima. Unlike a basic bowl-shaped surface, we anticipate encountering complexities, including bulges and diverse curvatures, making the 2D representation more intricate and varied than a simple bowl. However, if we consider adding any coefficient to the linear equation $h_{\theta}(x) = \theta_0 + \theta_1 x $, it won't transform into a non-linear equation. Now, picture this with a fixed loss, activation function, layers, dataset. We'll essentially be navigating around a somewhat related loss landscape. By employing Xavier Glorot initialization with a normal distribution, we end up landing in different areas of that loss region.

In this model setup, we may have either a convex or non-convex structure, but there will always be a point where the independent variables significantly influence or exhibit sensitivity to the dependent variables. Considering the gradients at a specific point in the space of multiple independent variables provides valuable information. Now, imagine having a history of these gradients throughout the process of minimizing the loss during backpropagation.

\section{Sensitivity of Independent Variables}
In the previous sections, we covered various aspects such as loss functions, activation functions, the shape of the loss surface, and how to set up a model that has been experimented and proven effective. Now, let's explore how gradients can help us understand the impact of a single feature on an independent variable. This concept is rooted in the first principle of calculus. Simply put, gradients provide insights into how changing one feature affects the outcome. It's like looking at the slope of a curve at a point.we gain a clearer understanding of their influence on the overall model. This process aligns with the fundamental principles of calculus, making it a powerful tool for analyzing sensitivity.

\subsection{Loss Landscape of Second Order Partial Derivative}
In the context of this model setup, where we only have a single column of second-order gradients  due to the limitations of calling it a Hessian matrix, now take a function $f(x)$ and its second-order partial derivative (\ref{secondOrder}), for a given function. This derivative provides insights into the curvature of the function with respect to the variable $x$. Specifically, a positive second derivative indicates a local minimum. if we are at a local minimum, the second-order gradient is expected to be positive. This is a crucial observation in understanding the behavior of the neural network during training.

\begin{equation}
\label{secondOrder}
\nabla^2 f = \frac{\partial^2 f}{\partial x^2} =\begin{bmatrix}
\frac{\partial^2 f}{\partial x_1^2}  \\
\frac{\partial^2 f}{\partial x_2 \partial x_1} \\
\vdots \\
\frac{\partial^2 f}{\partial x_n \partial x_1}&
\end{bmatrix}
\end{equation}

Another way to look at this is by observing the pattern of the second-order gradients. If these gradients are on the rise, it indicates that the features linked to that dimension are impactful and play a significant role in the minimization process. On the other hand, if the second-order gradients are on a downward trend during training, it suggests that the related features are not contributing effectively to the reduction of the loss. In simpler terms, these features may not be as vital for predicting the output or achieving a local minimum. Keeping an eye on the behavior of second-order gradients can thus help us in pinpointing the effect of specific features on the training process.

In essence, studying the second-order partial derivative offers crucial information about the curvature of the loss function in relation to a specific variable. This knowledge can be used to comprehend the importance of features and their contribution in achieving a local minimum during the training of neural networks.Furthermore, this understanding of the second-order gradients can be instrumental in refining the feature selection process. By identifying which features have a significant impact on the minimization process, we can prioritize these influential features in our model. This can lead to more efficient training and potentially improve the accuracy of our predictions.

Moreover, this analysis can also help in troubleshooting and optimizing the training process. If certain features are found to be not contributing effectively to the minimization of the loss, we can consider excluding these features or re-evaluating their representation in the model. This can help in reducing the complexity of the model and speed up the training process, without significantly affecting the performance of the model.

\subsection{Spearman correlation coefficient and Second order partial derivative}

The Spearman correlation coefficient assesses the strength and direction of the monotonic relationship between two variables, emphasizing the order of values rather than their specific numerical differences. It is not sensitive to non-linear relationships. A coefficient close to 1 indicates a strong positive correlation, meaning that as one variable increases, the other tends to increase as well. Conversely, a coefficient close to -1 signifies a strong negative correlation, suggesting that as one variable increases, the other tends to decrease. Thus, non-linearity is not taken into consideration by the Spearman. Our answers might be improved if we included a non-linear approach of how the independent variable influences the dependent variable. 

In Spearman correlation, the result remains the same for a given dataset. However, in this second order gradients, with retraining and exploration of the loss landscape, observing how the loss decreases becomes crucial. Specifically, when the loss decreases and the gradient decreases, it indicates that the dimension is not significantly contributing to the dependent variable. On the other hand, when the loss decreases and the gradient increases, it suggests that the respective dimension is influential or has an impact on the dependent variable.

\begin{figure}
    \centering
    \includegraphics[width=0.75\linewidth]{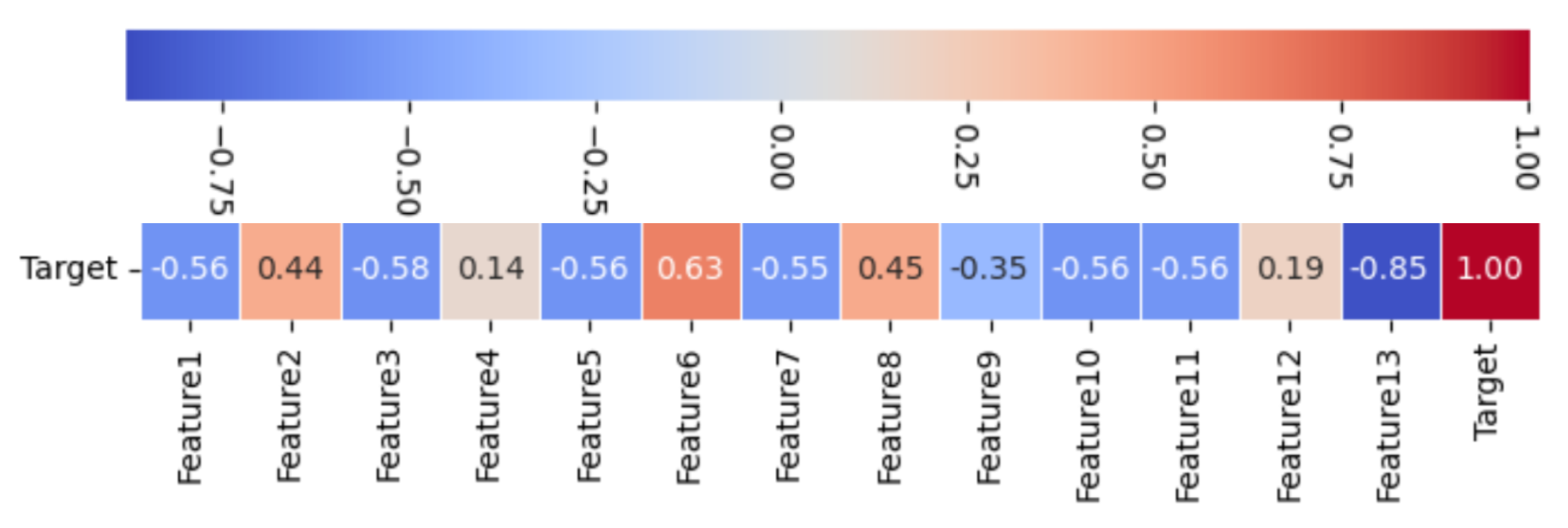}
    \caption{Spearman correlation coefficient}
    \label{fig:spearmanFig}
\end{figure}

\begin{figure}
    \centering
    \includegraphics[width=1\linewidth]{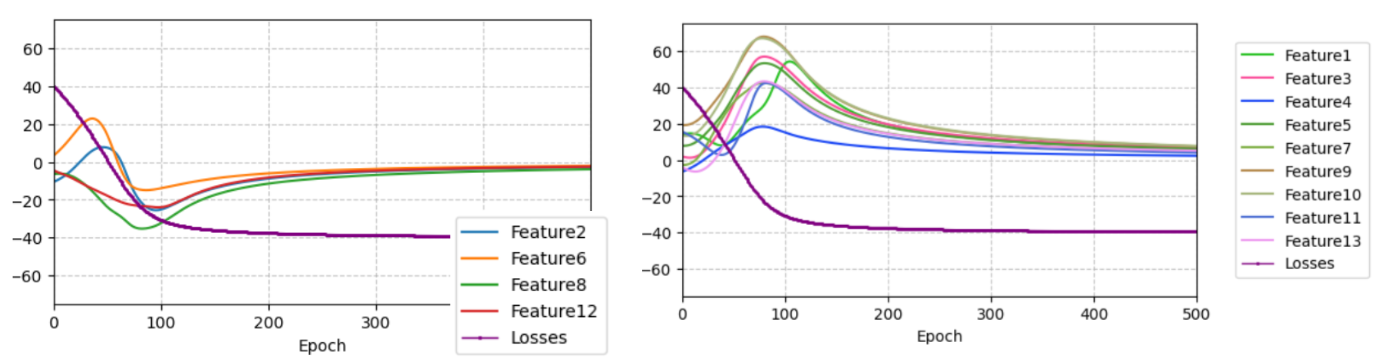}
    \caption{Second order gradient}
    \label{fig:secondGradient}
\end{figure}

Now, using the target as an independent variable, the Spearman correlation coefficient for a particular dataset with 13 columns is computed in Figure \ref{fig:spearmanFig}. For the same dataset, the history of the second order gradient of the training and losses is kept in Figure \ref{fig:secondGradient}.

\paragraph{New Insights on Second Order Gradients}

In Figure \ref{fig:secondGradient}, we observe that some feature's gradients are increasing, indicating an approach towards local minima, while others are decreasing, suggesting a movement away from local minima. Although gradients and Spearman correlation may seem unrelated coming from calculus and statistics, respectively through various experiments on different datasets and surfaces, I noticed some coincidences between Spearman correlation and the observed gradients. It's essential to note that these fields are distinct, yet there appears to be a connection.

Upon examining Figure \ref{fig:spearmanFig}: features with positive Spearman values also show an increase in Figure \ref{fig:secondGradient}, and features with negative Spearman values demonstrate a decrease as losses decreases. Through multiple experiments and retraining on diverse datasets, this pattern held consistently. Occasionally, there were minor fluctuations where features changed from increasing to decreasing or vice versa, but this was limited to only a few features, typically around 1 or 2 out of 13 for this dataset , What I found holds true for various datasets, however there are a few instances where the gradients are the only pointing to a larger number of correlation coefficients in the features.

\paragraph{Advantage of Second-Order Gradients than Spearman Correlation} In this specific model setup, unlike Spearman correlation, there's a hint of non-linearity introduced through the use of $MSE(\tanh(\theta_0 + \theta_1 x))$. As a result, it is similar to \textit{Non-linearity + Monotonic Relationship}. If you're interested in exploring beyond a simple Spearman monotonic relationship, this approach allows you to search deeper into the independent variables and their impact on the dependent variable. The goal isn't necessarily to extensively train the model for better accuracy; rather, it's about modestly reducing the loss and visually examining the relationships. If a more detailed perspective is needed, consider retraining the model with different parameters, but for now, focus on minimizing the loss and exploring insights through plotting.

\subsection{First and Third Order Partial Derivatives}
First order gradients are used to construct the second order gradient, while the second order gradient is instrumental in calculating the third order gradient. 

\begin{figure}
    \centering
    \includegraphics[width=0.50\linewidth]{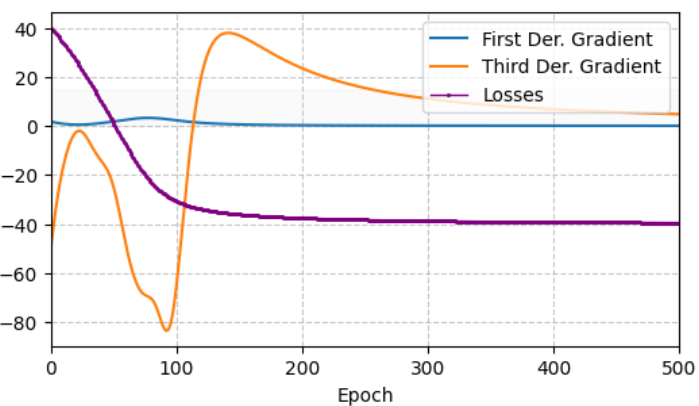}
    \caption{First and second order gradients}
    \label{fig:firstSecondfig}
\end{figure}

These gradients are straightforward because they both converge to zero, resulting in a local minimum. This property makes them particularly useful in optimization problems. The first and third order gradients for a single feature are depicted in Figure \ref{fig:firstSecondfig}. As can be observed, both gradients of this feature are progressively approaching zero as the loss decreases, indicating that this feature has a significant impact on the dependent variable.

\begin{equation}
\label{firstSecond}
\nabla f = \frac{\partial f}{\partial x}
;
\nabla^3 f = \frac{\partial^3 f}{\partial x^3}
\end{equation}

The first order gradient provides us with the rate of change of a function at a particular point. It essentially tells us the direction in which the function is increasing or decreasing. The third order gradient, on the other hand, gives us an insight into the rate of change of the second order gradient. 

In conclusion, understanding and utilizing first and third order gradients can be useful to know which independent variable influences the dependent variable. 
\subsection{Possibility of Gradients in images} 
Gradients in images can serve a purpose beyond sensitivity analysis, much like the human eye's blind spot, which allows us to focus on key details while disregarding the rest. This concept is particularly useful in real-time models where not all information is required. The human eye's blind spot is due to the absence of light-detecting cells where the optic nerve connects to the brain. Similarly, when we focus on a specific task, we often overlook other things happening around us. 

\textbf{Selective Attention}: We often miss things around us when focusing on a specific task. This is called selective attention. Gradients can be used for this with more research, which is vital in many applications. In the context of machine learning or AI models, it's often unnecessary to process all available information. Instead, these models can concentrate on the most relevant features. Thus, the use of gradients in images can significantly enhance the efficiency and effectiveness of these models.

\begin{figure}
    \centering
    \includegraphics[width=0.3\linewidth]{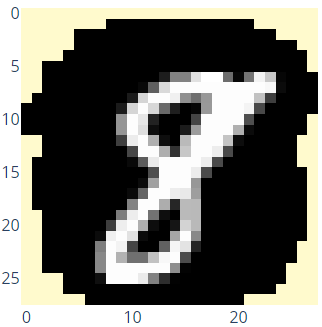}
    \caption{Image with gradients}
    \label{fig:mnistmask}
\end{figure}

I conducted a small experiment using the MNIST dataset. I trained a model using a subset of the dataset and stored the second order gradients. I then analyzed the second order gradient to identify the elements that increased and decreased. The decreasing elements were deemed not useful for prediction. From this, I created a single mask  using the gradients and applied it to the MNIST image Figure \ref{fig:mnistmask}. However, there are several limitations to this approach. For instance, the computational requirements are high and the MNIST dataset is not representative of all image datasets due to its fixed portion. Nonetheless, this was an experimental thought and a step towards understanding the potential of gradients in image processing.

\section{Conclusion}
In conclusion, this study presents a novel approach to sensitivity analysis by exploring the loss landscape through the use of first, second, and third derivatives. The findings reveal that second-order gradients, under certain configurations, can provide insights similar to Spearman's correlation coefficient, but with the added advantage of detecting non-linear relationships. This approach allows for a deeper understanding of the impact of independent variables on the dependent variable. While the focus is not on extensive model training for accuracy, the method provides valuable insights through the visualization of relationships. Future work could involve retraining the model with different parameters for a more detailed perspective. This study underscores the importance of sensitivity analysis in enhancing model robustness and providing a comprehensive understanding of the loss landscape.

\bibliographystyle{unsrt}  
\bibliography{references}  

\end{document}